# Optimal Tracking Controller Design for a Small Scale Helicopter

**Agus Budiyono\*** and **Singgih S. Wibowo**=

\*Aeronautics and Astronautics Department
Institut Teknologi Bandung, Indonesia
E-mail: agus.budiyono@ae.itb.ac.id

=Aerospace and Software Engineer
Simundo – a Simulation Technology Company
Bandung, Indonesia
E-mail: singgih_wibowo@yahoo.com

**Abstract**

A model helicopter is more difficult to control than its full scale counterparts. This is due to its greater sensitivity to control inputs and disturbances as well as higher bandwidth of dynamics. This works is focused on designing practical tracking controller for a small scale helicopter following predefined trajectories. A tracking controller based on optimal control theory is synthesized as part of the development of an autonomous helicopter. Some issues in regards to control constraints are addressed. The weighting between state tracking performance and control power expenditure is analyzed. Overall performance of the control design is evaluated based on its time domain histories of trajectories as well as control inputs.

## 1 Introduction

The period of 1990s has witnessed the pervasive use of classical control systems for a small scale helicopter [1]. A single-input-single-output SISO proportional-derivative (PD) feedback control systems have been primarily used in which their controller parameters were usually tuned empirically. The approach is based on performance measures defined in the frequency and time domain such as gain and phase margin, bandwidth, peak overshoot, rise time and settling time. This trial-and-error approach to design an "acceptable" control system however is not agreeable with a more general complex multi-input multi-output MIMO systems with sophisticated performance criteria. To control a model helicopter as a complex MIMO system, an approach that can synthesize a control algorithm to make the helicopter meet performance criteria while satisfying some physical constraints is required. Recent developments in this line of research include the use of optimal control (Linear Quadratic Regulator) implemented on a small aerobatic helicopter designed at MIT [2,3]. Similar approach has been also independently developed for a rotor unmanned aerial vehicle at UC Berkeley [1]. An adaptive high-bandwidth controller for helicopter was synthesized at Georgia Technology Research Institute [4].

The current paper addresses this challenge using optimal control theory and reports encouraging preliminary results amenable to its application to multivariable control synthesis for high bandwidth dynamics of a small scale helicopter. The practical tracking controller is intended to be implemented on the on-board computer of the model helicopter as part of its autonomous system design.

## 2 Dynamics of a Small Scale Helicopter

The Yamaha R-50 helicopter dynamics model has been developed at Carnegie Mellon Robotics Institute. The experimental helicopter is shown in Fig. 1. It uses a two bladed main rotor with a Bell-Hiller stabilizer bar. The physical characteristic of the helicopter is summarized in Table 1. The basic linearized equations of motion for a model helicopter dynamics are derived from the Newton-Euler equations for a rigid body that has six degrees of freedom to move in space. The external forces, consisting aerodynamic and gravitational forces, are represented in a stability derivative form. For simplicity, the control forces produced by the main and tail rotor are expressed by the multiplication of a control derivative and the associated control input. Following [5], the equations of motion of the model helicopter are derived and categorized into the following groups.

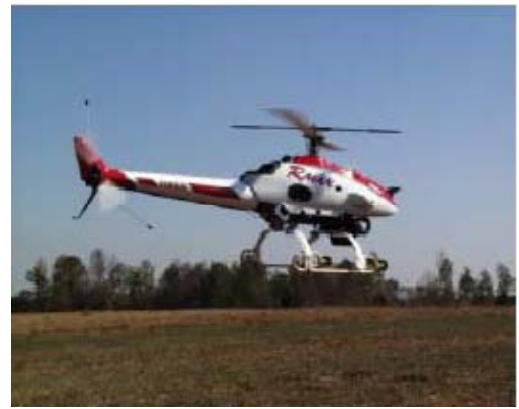

**Figure 1:** The experimental R-50 helicopter

**Table 1: Physical Parameter of the Yamaha R-50**

| | |
|---|---|
| Rotor speed | 850 rpm |
| Tip speed | 449 ft/s |
| Dry weight | 97 lb |
| Instrumented | 150 lb |
| Engine | Single cylinder, 2-stroke |
| Flight autonomy | 30 minutes |

### 2.1 Lateral and longitudinal fuselage dynamics

Using the Newton-Euler equations, the translational and angular fuselage motions of the helicopter can be derived as the following set of equations:

$$\dot{u} = (w_0 q + v_0 r) - g\theta + X_u u + X_a a \quad (1)$$
$$\dot{v} = (-u_0 r + w_0 p) - g\phi + Y_v v + Y_b b \quad (2)$$
$$\dot{p} = L_u u + L_v v + \cdots + L_b b \quad (3)$$
$$\dot{q} = M_u u + M_v v + \cdots + M_a a \quad (4)$$

The stability derivatives are used to express the external aerodynamic and gravitational forces and moments. $X_a$, $Y_b$ denotes rotor derivatives and $L_b$, $M_a$ the flapping spring-derivatives. They are all used to describe the rotor forces and moments respectively. General aerodynamic effects are expressed by speed derivatives given as $X_u$, $Y_v$, $L_u$, $L_v$, $M_u$, $M_v$.

### 2.2 Heaving (vertical) dynamics

The Newton-Euler rigid body equations for the heaving dynamics is represented by

$$\dot{w} = (-v_0 p + u_0 q) + Z_w w + Z_{col} \delta_{col} \quad (5)$$

In the hovering flight, $v_0$ and $u_0$ are obviously zero. Thus, the centrifugal forces represented by the terms in parentheses are relevant only in cruise flight.

### 2.3 Yaw dynamics

The augmented yaw dynamics is approximated as a first order bare airframe dynamics with a yaw rate feedback represented by a simple first-order low-pass filter. The corresponding differential equations used in the state-space model are also given in appropriate stability derivatives as follows

$$\dot{r} = N_r r + N_{ped}(\delta_{ped} - r_{fb}) \quad (6)$$
$$\dot{r}_{fb} = -K_{rfb} r_{fb} + K_r r \quad (7)$$

### 2.4 Coupled Rotor-stabilizer bar dynamics

The simplified rotor dynamics is represented by two first-order differential equations for the lateral ($b$) and longitudinal ($a$) flapping motion. In the state-space model, the rotor models are given as:

$$\dot{b} = \frac{-b - \tau_f p + B_a a + B_{lat}(\delta_{lat} + K_d d) + B_{lon}\delta_{lon}}{\tau_f} \quad (8)$$
$$\dot{a} = \frac{-a - \tau_f q + A_b b + A_{lat}\delta_{lat} + A_{lon}(\delta_{lon} + K_c c)}{\tau_f} \quad (9)$$

where the following derivatives related to the gearing of the Bell-mixer are introduced:

$$K_d = \frac{B_d}{B_{lat}} \quad (10)$$
$$K_c = \frac{A_c}{A_{lon}} \quad (11)$$

The stabilizer bar receives cyclic inputs from the swashplate in a similar way as do the main blades. The equations for the lateral ($d$) and longitudinal ($c$) flapping motions are:

$$\dot{d} = \frac{-d - \tau_s p + D_{lat}\delta_{lat}}{\tau_s} \quad (12)$$
$$\dot{c} = \frac{-c - \tau_s q + C_{lon}\delta_{lon}}{\tau_s} \quad (13)$$

### 2.5 The state-space model of the R-50 dynamics

The state-space model of the helicopter can be assembled from the above set of differential equations in a matrix form:

$$\underline{\dot{x}} = A\underline{x} + B\underline{u} \quad (14)$$

where $\underline{x} = \{u, v, p, q, \varphi, \theta, a, b, w, r, r_{fb}, c, d\}^T$ is the state vector and $\underline{u} = \{\delta_{lat}, \delta_{lon}, \delta_{ped}, \delta_{col}\}^T$ the input vector. The dynamic matrix $A$ contains the stability derivatives and the control matrix $B$ contains the input derivatives. The complete description the elements of these matrices are presented in the Appendix.

## 3 Tracking Control Design

### 3.1 Linear Regulator Problem

A Linear Regulator problem in the optimal control theory represents a class of problem where the plant dynamics is linear and the quadratic form of performance criteria is used. The linear dynamics (which can be time-varying) are:

$$\underline{\dot{x}}(t) = A(t)\underline{x}(t) + B(t)\underline{u}(t) \quad (15)$$

and the cost is quadratic:

$$J = \tfrac{1}{2}\underline{x}(t_f)^T H \underline{x}(t_f)$$
$$+ \tfrac{1}{2}\int_{t_0}^{t_f} \int [\underline{x}(t)^T Q(t)\underline{x}(t) + \underline{u}(t)^T R(t)\underline{u}(t)]dt \quad (16)$$

where the requirements of the weighting matrices are given as

$$H = H^T \geq 0 \quad (17)$$
$$Q(t) = Q(t)^T \geq 0$$
$$R(t) = R(t)^T \geq 0 \quad (18)$$

Also, there is no other constraints and $\tau_f$ is fixed i.e. no terminal constraints and no constraints on $\underline{u}$. Note that there is a terminal weighting of the states $\underline{x}$. The physical interpretation of the problem statement is: it is desired to maintain the state vector close to the origin without an excessive expenditure of control effort [6]. The optimal feedback control law can be derived by identifying the Hamiltonian of the system and using the Hamilton-Jacoby-Bellman (HJB) equation to ensure the optimality [7]. The Hamiltonian $H = H[\underline{x}(t), \underline{u}^*(t), J_x^*, t]$ for the above problem is defined as:

$$H = \tfrac{1}{2}\underline{x}(t)^T Q(t)\underline{x}(t) + \tfrac{1}{2}\underline{u}(t)^T R(t)\underline{u}(t) \\ + J_x^*(\underline{x}, t) \cdot [A(t)\underline{x}(t) + B(t)\underline{u}(t)] \quad (19)$$

minimizing with respect to $\underline{u}$, it reads:

$$\tfrac{dH}{d\underline{u}} = \underline{u}(t)R(t) +^T J_x^*(\underline{x}, t) \cdot B(t) \\ \tfrac{d^2H}{d\underline{u}} = R(t) > 0 \quad (20)$$

Note that the above stationary condition defines a global minimum because the function $H$ depends quadratically only on $\underline{u}$.

The optimal control is obtained by using the stationary condition and solving for $\underline{u}$,

$$\underline{u}^*(t) = -R(t)^{-1} B(t)^T J_x^*(\underline{x}, t)^T \quad (21)$$

The Hamiltonian in Eq. (19) can then be written as:

$$H = J_x^*(\underline{x}, t) \cdot A(t)\underline{x}(t) + \tfrac{1}{2}\underline{x}(t)^T Q(t)\underline{x}(t) \\ -\tfrac{1}{2}J_x^*(\underline{x}, t) \cdot B(t)R(t)^{-1}B(t)^T J_x^*(\underline{x}, t)^T \quad (22)$$

Now, in order to ensure the optimality the HJB equation must be satisfied. The complete HJB equation is:

$$J_t^*(\underline{x}, t) = J_x^*(\underline{x}, t)^T \cdot A(t)\underline{x}(t) + \tfrac{1}{2}\underline{x}(t)^T Q(t)\underline{x}(t) \\ -\tfrac{1}{2}J_x^*(\underline{x}, t) \cdot B(t)R(t)^{-1}B(t)^T J_x^*(\underline{x}, t)^T \quad (23)$$

where the boundary condition (in this case, with fixed $t_f$) is

$$J^*(\underline{x}(t_f), t_f) = \tfrac{1}{2}\underline{x}(t_f)^T H \underline{x}(t_f) \quad (24)$$

A quadratic form for the cost is used to verify its validity using the HJB equation. Letting:

$$J^*(\underline{x}(t), t) = \tfrac{1}{2}\underline{x}(t)^T H \underline{x}(t) \quad (25)$$

Thus, the partial derivatives appearing in Eq. (23), can be written as

$$J_x^*(\underline{x}, t) = \underline{x}(t)^T K(t) \quad (26)$$
$$J_t^*(\underline{x}, t) = \tfrac{1}{2}\underline{x}(t)^T \tfrac{dK(t)}{dt} \underline{x}(t) \quad (27)$$

and the HJB equation becomes:

$$\tfrac{1}{2}\underline{x}(t)^T \dot{K}(t)\underline{x}(t) = \tfrac{1}{2}\underline{x}(t)^T Q(t)\underline{x}(t) + \underline{x}(t)^T K(t)A(t)\underline{x}(t) \\ -\tfrac{1}{2}\underline{x}(t)^T K(t)B(t)\underline{x}(t)R(t)^{-1}B(t)^T K(t)^T \underline{x}(t) \quad (28)$$

The terms appearing in the above equation are scalar. The transpose of a scalar term is the same as the term. Since only scalar terms are dealt with the following relation applies:

$$\underline{x}(t)^T K(t)A(t)\underline{x}(t) = \tfrac{1}{2}\underline{x}(t)^T K(t)A(t)\underline{x}(t) \\ + \tfrac{1}{2}\underline{x}(t)^T A(t)^T K(t)^T \underline{x}(t) \quad (29)$$

And thus the HJB equation becomes:

$$0 = \tfrac{1}{2}\underline{x}(t)^T [\dot{K}(t) + Q(t) + K(t)A(t) + A(t)^T K(t)^T \\ - K(t)B(t)R(t)^{-1}B(t)^T K(t)^T]\underline{x}(t) \quad (30)$$

Now, letting $K(t_f) = H = H^T$ (symmetric), $K(t)$ will be symmetric from above and thus symmetry of $K(t_f)$ will be retained $\forall t$, i.e. $K(t) = K(t)^T$ (symmetric). $K(t_f) = H$ is the boundary condition for $K(t)$. Since the above equation must be satisfied $\forall \underline{x}(t)$, the following matrix differential equation is obtained:

$$-\dot{K}(t) = Q(t) + K(t)A(t) + A(t)^T K(t)^T \\ - K(t)B(t)R(t)^{-1}B(t)^T K(t)^T \quad (31)$$

With a finite terminal time specified the entire solution is a transient and $K(t)$ will be time varying. However if the terminal time is taken far enough out, the solution for $K$ and the corresponding feedback gain might tend to a constant. To get the invariant asymptotic solution the differential equation for $\dot{K}$ can be integrated to a steady state solution or the time derivative term is set to zero i.e. $\dot{K} = 0$. This leads to Algebraic Riccati equation (ARE) [6]:

$$Q + KA(t) + A^T K^T - KBR^{-1}B^T K^T = 0 \quad (32)$$

For $K(t)$ satisfying the above Riccati form matrix quadratic equation, it is required that Eq.(25) is optimal, and the optimal control in Eq.(21) is now given by

$$\underline{u}^*(t) = -R^{-1}B^T K \underline{x}(t) \quad (33)$$

This is the state feedback control law for the continuous time LQR problem. Note that both forms of Riccati equation given in Eqs. (31) and (32) do not depend on $\underline{x}$ or $\underline{u}$ and thus the solution for $K$ can be obtained in advance and finally the gain matrix $R^{-1}B^T K$ can be stored. The control can then be obtained in real time by multiplying the stored gain with $\underline{x}(t)$.

### 3.2 Path tracking problem formulation

To use the LQR design for path tracking control, the regulator problem must be recast as a tracking problem. In a tracking problem, the output y is compared to a reference signal r. The goal is to drive the error between the reference and the output to zero. It is common to add an integrator to the error signal and then minimize it. An alternative approach would be using the derivative of the error signal. Assuming perfect measurements i.e. the sensor matrix is of identity form:

$$\underline{y}_{error} = \underline{x}_{error}(t) = \underline{x}_{ref}(t) - \underline{x}(t) \quad (34)$$

Taking the time derivative of the equation yields:

$$\underline{\dot{x}}_{error}(t) = \underline{\dot{x}}_{ref}(t) - \underline{\dot{x}}(t) \quad (35)$$

When the reference is predefined as constant, then $\underline{\dot{x}}_{ref}(t) = 0$, and

$$\underline{\dot{x}}_{error}(t) = -\underline{\dot{x}}(t) \quad (36)$$

A path tracking control law can be designed by using the following general relation:

$$\underline{\dot{x}}_{error}(t) = -\eta \underline{\dot{x}}(t) \quad (37)$$

where $\eta$ is an arbitrary constant representing the weight of the tracking performance in the cost function. In the matrix form, the above relation can be written as:

$$\underline{\dot{x}}_{error}(t) = \begin{bmatrix} \dot{x}_{error}(t) \\ \dot{y}_{error}(t) \\ \dot{z}_{error}(t) \end{bmatrix} = \begin{bmatrix} -\eta \dot{x}(t) \\ -\eta \dot{y}(t) \\ -\eta \dot{z}(t) \end{bmatrix} \quad (38)$$

Substituting $\dot{x} = u$, $\dot{y} = v$, $\dot{z} = w$, the above equation can be expressed as

$$\underline{\dot{x}}_{error}(t) = \begin{bmatrix} -\eta u(t) \\ -\eta v(t) \\ -\eta w(t) \end{bmatrix} \quad (39)$$

To accommodate the tracking term in the cost function, the state-space model is augmented as the following:

$$\underline{\dot{x}}_{aug} = A_{aug}\underline{x}_{aug}(t) + B_{aug}\underline{u}(t) \quad (40)$$

where:

$$\underline{x}_{aug} = \{\underline{x}_{error}, \underline{x}\}^T \quad (41)$$

$$A_{aug} = \begin{bmatrix} 0_{3\times3} & -\eta I_3 & 0_{3\times11} \\ 0_{14\times3} & & 0_{3\times3} \end{bmatrix} \quad (42)$$

$$B_{aug} = \begin{bmatrix} 0_{3\times4} \\ B \end{bmatrix} \quad (43)$$

The size of the matrix is associated with the matrix appearing in Eq.(14) with the addition of three augmented states. When the terminal weighting is not considered, the performance measure is now

$$J = \tfrac{1}{2}\int_{t_0}^{t_f}\int \left[\underline{x}_{aug}(t)^T Q(t)\underline{x}_{aug}(t) + \underline{u}(t)^T R(t)\underline{u}(t)\right]dt \quad (44)$$

where the determination of the weighting matrices $\eta$, $Q$ and $R$ are empirical.

### 3.3 Control synthesis in the presence of input constraints

In a real system, the control inputs are always limited by hard constraints. The limitation of the control inputs for a typical aircraft, for instance, is governed by the maximum allowable deflection of its control surfaces. In the case of helicopters, the control hard limits are imposed on their lateral and longitudinal cyclic, pedal and collective pitch. The control input limits for the helicopter used in this study are:

$$-5^o \le \delta_{lat} \le 5^o \quad (45)$$
$$-5^o \le \delta_{lon} \le 5^o \quad (46)$$
$$-22^o \le \delta_{ped} \le 22^o \quad (47)$$
$$-10^o \le \delta_{col} \le 10^o \quad (48)$$

To incorporate control input constraints into the control synthesis, the Pontryagin's minimum principle is employed. Essentially, the Hamiltonian of the system is re-derived to express the presence of input constraints. The stationary condition is applied to the modified Hamiltonian to obtained the optimal bounded control [8].

The Hamiltonian for the above tracking problem is

$$H = \tfrac{1}{2}\underline{x}_{aug}(t)^T Q(t)\underline{x}_{aug}(t) + \tfrac{1}{2}\underline{u}(t)^T R(t)\underline{u}(t)$$
$$+p^T\left(A\underline{x}_{aug}(t) + B\underline{u}(t)\right) \quad (49)$$

and its derivative with respect to $\underline{u}$ is

$$H_u = \underline{u}(t)^T R(t) + p^T B \quad (50)$$

The optimal control can be solved by imposing $H_u = 0$. This yield

$$\underline{u}^*(t) = -R(t)^{-1}B^T \underline{p} \quad (51)$$

When the control is constrained or bounded, the optimal control is

$$\underline{u}^*(t) = \underset{\underline{u}(t) \in U(t)}{\arg\min}\left[\underline{u}(t)^T R(t)\underline{u}(t) + p^T B\underline{u}(t)\right] \quad (52)$$

In practice, the control elements are penalized individually. It does not make any sense to minimize the product between $u_1$ and $u_2$, for example. Thus the weighting matrix $R$ is not usually a full matrix. The diagonal matrix $R$ was used in this work which simplified the optimization considerably. For a diagonal $R$,

$$\underline{u}^*(t) = \underset{\underline{u}(t) \in U(t)}{\arg\min}\left[\sum_{i=1}^{m}\tfrac{1}{2}R_{ii}u_i^2 + p^T b_i u_i\right] \quad (53)$$

$$u_i^*(t) = \underset{u_i(t) \in U_i(t)}{\arg\min}\left[\tfrac{1}{2}R_{ii}u_i^2 + p^T b_i u_i\right] \quad (54)$$

The unbounded solution can thus be written as

$$\tilde{u}_i = -R_{ii}^{-1}p^T b_i \quad (55)$$

and the bounded control requirement, $-M_i^- \leq u_i \leq M_i^+$, can be implemented in the following logic:

$$\begin{array}{ll} if\ \tilde{u}_i \leq -M_i^- & u_i^* = -M_i^- \\ -M_i^- \leq \tilde{u}_i \leq M_i^+ & u_i^* = \tilde{u}_i \\ \tilde{u}_i \geq M_i^+ & u_i^* = M_i^+ \end{array} \quad (56)$$

Note that the solution of the bounded control is not the same as the solution obtained by imposing the constraints to the unbounded solution. The optimal control history of the bounded control case cannot be determined by calculating the optimal control history for the unbounded case and then allowing it to saturate whenever there is a violation of the stipulated boundaries.

## 4 Numerical Simulation

To evaluate the performance of the tracking controller design, numerical simulation is conducted for a variety of reference trajectories and for different values of weighting matrices. The simulation is carried out using Matlab with the data presented in the Appendix. Since the focus of the design is on the tracking performance, only the data corresponding to cruise is applicable. The effects of the weighting matrices are analyzed based on the evident tradeoff between tracking performance and control expenditure. The effect of the weighting matrix to the states will be observed by comparing the tracking performance between two highly separated values of Q, $Q = 0.01I_{17}$ and $Q = I_{17}$. Similar effects will be observed for position tracking and control weighting matrices.

Note that in order to be able to physically interpret the result of the simulation, a coordinate transformation is needed between body coordinate and local horizon coordinate system. The transformation matrix between these two coordinates is given as the following:

$$T_l^b = \begin{bmatrix} c\theta c\psi & c\theta s\psi & -s\theta \\ s\phi s\theta c\psi - c\phi s\psi & s\phi s\theta s\psi + c\phi c\psi & s\phi c\theta \\ c\phi s\theta c\psi + s\phi s\psi & c\phi s\theta s\psi - s\phi c\psi & c\phi c\theta \end{bmatrix} \quad (57)$$

where $c(\cdot) = \cos(\cdot)$, and $s(\cdot) = \sin(\cdot)$

With this transformation, the final results of the simulation are presented in the local horizon coordinate. The corresponding equations for the positions and the velocities are:

$$[N, E, A]^T = T_l^b [x, y, z]^T$$
$$[V_x, V_y, V_z]^T = T_l^b [u, v, w]^T \quad (58)$$

*Case 1*: Reference: rectangular trajectory as given in Fig.[2]. The weighting matrices are $\eta = 0.01$, $Q = 0.01I_{17}$, $R = 0.01I_4$.
*Case 2*: Reference: rectangular continued by circular trajectory as given in Fig.[5]. The weighting matrices are $\eta = 0.01$, $Q = 0.01I_{17}$, $R = 0.01I_4$.
*Case 3*: Reference: rectangular trajectory as given in Fig.[8]. The weighting matrices are $\eta = 5, Q = I_{17}$, $R = I_4$.
*Case 4*: Reference: rectangular continued by circular trajectory as given in Fig.[11]. The weighting matrices are $\eta = 5, Q = I_{17}$, $R = I_4$.

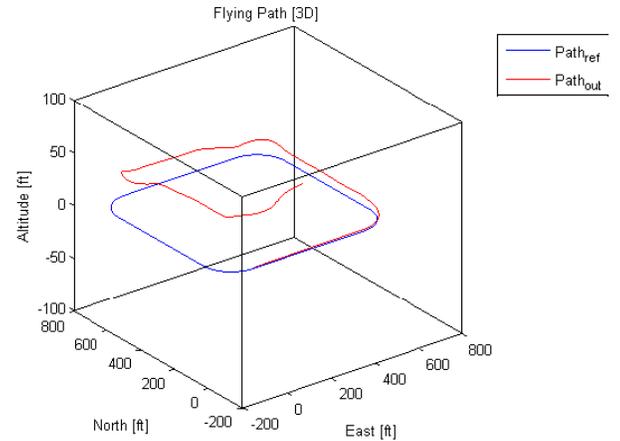

**Figure 2:** Trajectory tracking performance, $\eta = 0.01$, $Q = 0.01I_{17}$, $R = 0.01I_4$

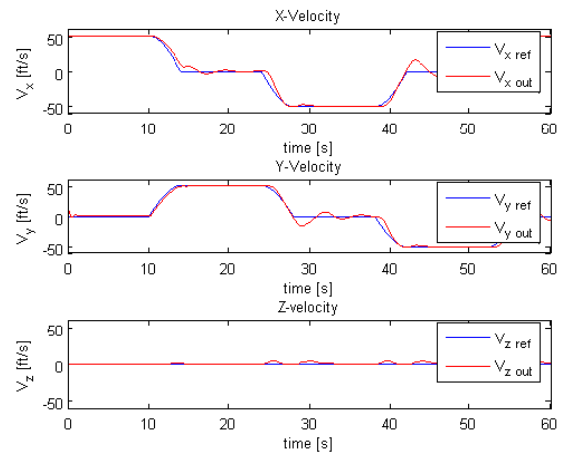

**Figure 3:** Velocity tracking performance, $\eta = 0.01$, $Q = 0.01I_{17}$, $R = 0.01I_4$

95

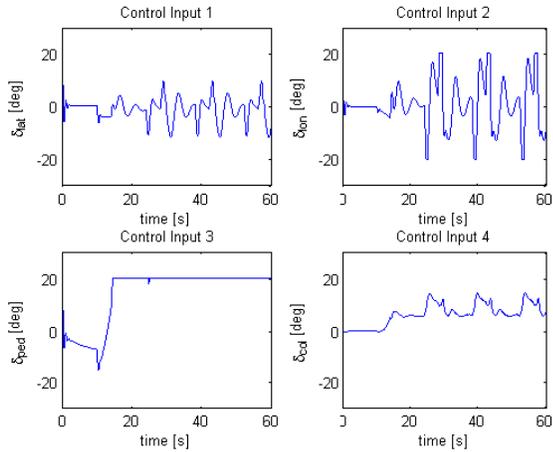

**Figure 4:** Control input expenditure, $\eta = 0.01$, $Q = 0.01I_{17}$, $R = 0.01I_4$

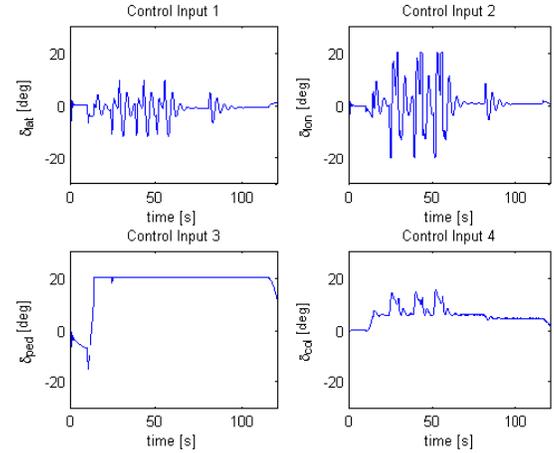

**Figure 7:** Control input expenditure, $\eta = 0.01$, $Q = 0.01I_{17}$, $R = 0.01I_4$

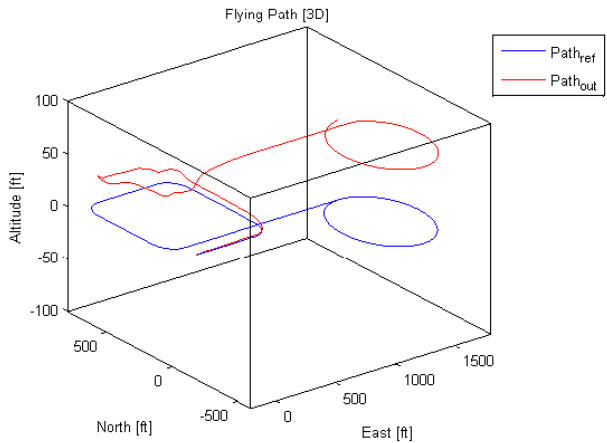

**Figure 5:** Trajectory tracking, $\eta = 0.01$, $Q = 0.01I_{17}$, $R = 0.01I_4$

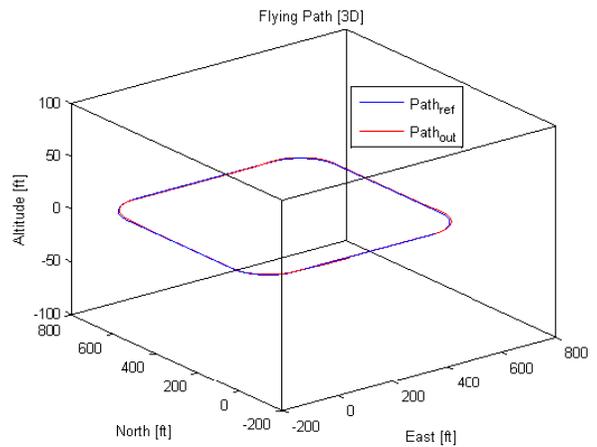

**Figure 8:** Trajectory tracking, $\eta = 5$, $Q = I_{17}$, $R = I_4$

## 5 Discussion on results

Fig.2 indicates the tracking performance for Case 1, where the diagonal elements of weighting matrices were assigned small numerical values. In the calculation of the optimal control, no constraints of control input were imposed. It can be observed that towards the end of the trajectory the tracking performance is degrading especially in the vertical position. The control input history, in Fig.4,
shows that even though the control expenditure is not bounded, it failed to provide acceptable tracking performance. The degradation in the tracking performance is more pronounced for a more complex reference trajectory in Case 2, shown in Fig.5.

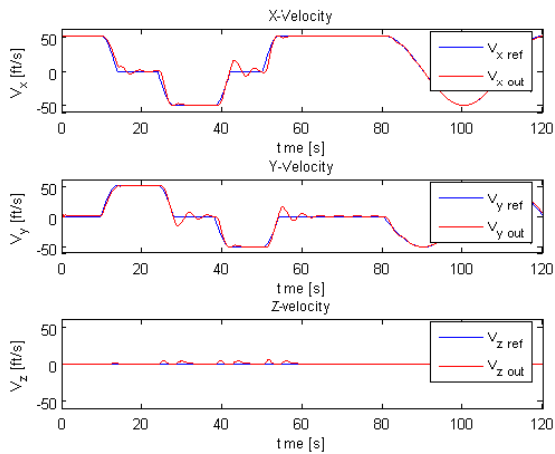

**Figure 6:** Velocity tracking, $\eta = 0.01$, $Q = 0.01I_{17}$, $R = 0.01I_4$

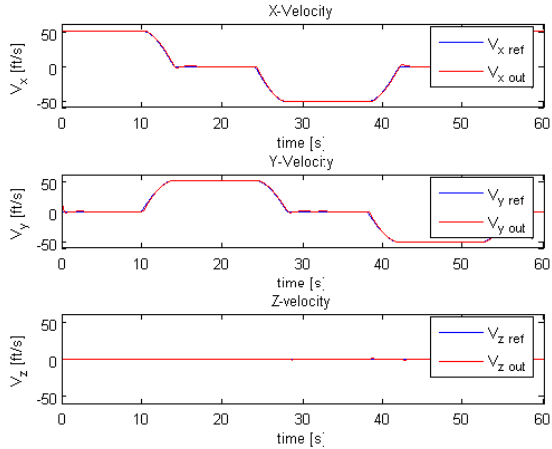

**Figure 9:** Velocity tracking performance, $\eta = 5$, $Q = I_{17}$, $R = I_4$

In Case 3 and 4, the weights for tracking and control elements were increased in order of magnitudes to observe their effect to the overall performance of tracking controller. Figs.8 through 10 show the tracking performance and the associated bounded control input for Case 3. The same controller can successfully handle more complex trajectory for Case 4 as shown in Fig.11 through 13. The tracking errors can be kept minimum while maintaining the required control input within the stipulated boundaries.

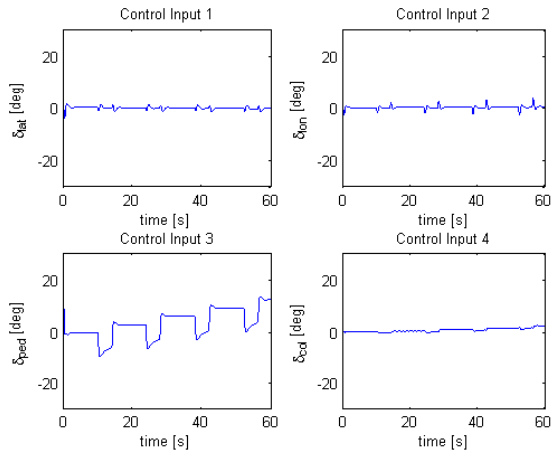

**Figure 10:** Control input expenditure, $\eta = 5$, $Q = I_{17}$, $R = I_4$

The numerical comparison between Case 1 and Case 3, shown in Table 2, represents the tracking performance comparison to evaluate the effect of the weighting matrices. The performance norm is given as the mean square error (MSE) between the actual and reference trajectories. It is evident that Case 3 ($Q = I17$) demonstrates order of magnitudes lower tracking error compared to Case 1 (with $Q = 0.01I17$). The effect of the weighting matrix Q is more pronounced for the position tracking of E and A. The MSE type tracking error for higher Q (Case 3) is 0.0142 and 0.000233 of that of the low value of Q (Case 1).

**Table 2: Tracking error comparison**

| | *Tracking error (MSE)* | | | | | |
|---|---|---|---|---|---|---|
| | Velocity(fps) | | | Position (ft) | | |
| Case | $V_x$ | $V_y$ | $V_z$ | N | E | A |
| 1 | 19.8 | 20.2 | 3 | 148 | 2043 | 771 |
| 3 | 0.77 | 1.13 | 0.06 | 9.6 | 29 | 0.18 |

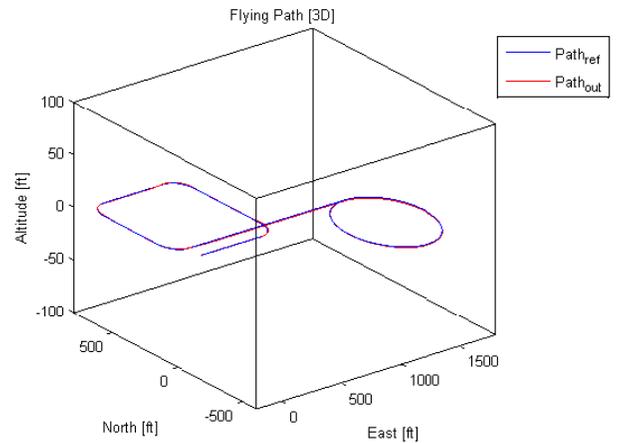

**Figure 11:** Trajectory tracking performance, $\eta = 5$, $Q = I_{17}$, $R = I_4$

Note that in this case study, the weight for augmented states and control element were assigned the same values. This assumption simply means that in the optimization process the control input elements are considered equally important. However, for the augmented states, the assumption means that the importance of the states is not treated evenly.

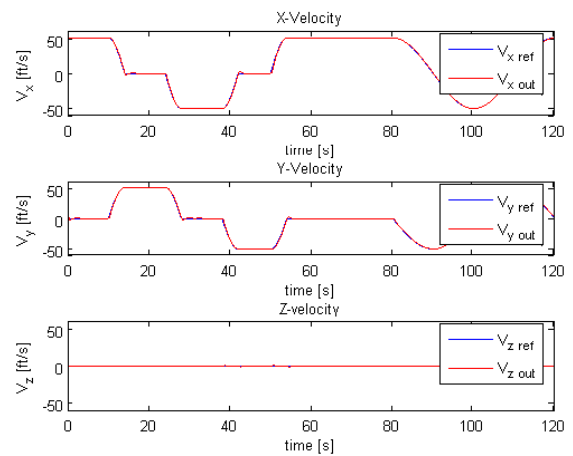

**Figure 12:** Velocity tracking performance, $\eta = 5$, $Q = I_{17}$, $R = I_4$

Overall results indicate that weight assignments play a significant role in the optimization process which determines the tracking performance. To the author's knowledge, despite its great influence, the determination of weighting values has been so far done primarily on an ad-hoc and case-by-case basis. The determination of the weighting values for the tracking performance of a ballistic missile involving its range and azimuth angle, for instance, can be guided by a simple fact that 0.01 radian error in the azimuth angle can contribute a position error of 10 km for missiles with a range of 1000 km. More refined weighting scheme using similar approach can be applied for the optimal control design of helicopters. It is worthy to note that formal treatment of weighting assignment for the optimal control methodology exists in the literature. One can use the pole placement technique in conjunction with optimal control theory where poles of the closed loop system were assigned and the weight assignment can be derived from the corresponding mathematical relation. A more novel technique has been recently proposed in Ref.[9] in the framework of polynomial approach where formal weighting assignment can be performed in association with integrated control design criteria.

## 6    Concluding Remarks

A control design methodology based on optimal control theory was elaborated and applied for the controller of a small scale helicopter model. It has been demonstrated that the approach neatly handled more complex design criteria than ones that can be traditionally afforded by classical control design. The overall design is part of an ongoing research, design and integration of a small autonomous helicopter [10] where robust yet practical control algorithm is desired. The anticipated practical control design criteria include, but are not limited to:

1. To fly the helicopter from an arbitrary origin to a specified waypoint in minimum time which characterizes a minimum-time problem
2. To bring the helicopter from an arbitrary initial state to a specified waypoint, with a minimum expenditure of control effort which is a minimum-control-effort problem
3. To minimize the deviation of the final state of the helicopter from its desired waypoint which represents a terminal control problem

The above design criteria can be conveniently formulated and incorporated into the cost function. For future research direction, it will be interesting to explore if the proposed control technique can maintain acceptable performance in the presence of wind.

**Appendix**

## I.  The state-space model of R-50 helicopter

The state-space equation describing the R-50 dynamics is:

$$\begin{bmatrix} \dot{u} \\ \dot{v} \\ \dot{p} \\ \dot{q} \\ \dot{\phi} \\ \dot{\theta} \\ \tau_f \dot{a} \\ \tau_f \dot{b} \\ \dot{w} \\ \dot{r} \\ \dot{r}_{fb} \\ \tau_s \dot{c} \\ \tau_s \dot{d} \end{bmatrix} = \begin{bmatrix} X_u & 0 & 0 & 0 & 0 & -g & X_a & 0 & 0 & 0 & 0 & 0 & 0 \\ 0 & Y_v & 0 & 0 & g & 0 & 0 & Y_b & 0 & 0 & 0 & 0 & 0 \\ L_u & L_v & 0 & 0 & 0 & 0 & 0 & L_b & L_w & 0 & 0 & 0 & 0 \\ M_u & M_v & 0 & 0 & 0 & 0 & M_a & 0 & M_w & 0 & 0 & 0 & 0 \\ 0 & 0 & 1 & 0 & 0 & 0 & 0 & 0 & 0 & 0 & 0 & 0 & 0 \\ 0 & 0 & 0 & 1 & 0 & 0 & 0 & 0 & 0 & 0 & 0 & 0 & 0 \\ 0 & 0 & 0 & -\tau_f & 0 & 0 & -1 & A_b & 0 & 0 & 0 & A_c & 0 \\ 0 & 0 & -\tau_f & 0 & 0 & 0 & B_a & -1 & 0 & 0 & 0 & 0 & B_d \\ 0 & 0 & 0 & 0 & 0 & 0 & Z_a & Z_b & Z_w & Z_r & 0 & 0 & 0 \\ 0 & N_v & N_p & 0 & 0 & 0 & 0 & 0 & N_w & N_r & N_{rfb} & 0 & 0 \\ 0 & 0 & 0 & 0 & 0 & 0 & 0 & 0 & 0 & K_r & K_{rfb} & 0 & 0 \\ 0 & 0 & 0 & -\tau_s & 0 & 0 & 0 & 0 & 0 & 0 & 0 & -1 & 0 \\ 0 & 0 & -\tau_s & 0 & 0 & 0 & 0 & 0 & 0 & 0 & 0 & 0 & -1 \end{bmatrix} \begin{bmatrix} u \\ v \\ p \\ q \\ \phi \\ \theta \\ a \\ b \\ w \\ r \\ r_{fb} \\ c \\ d \end{bmatrix} + \begin{bmatrix} 0 & 0 & 0 & 0 \\ 0 & 0 & Y_{ped} & 0 \\ 0 & 0 & 0 & 0 \\ 0 & 0 & 0 & M_{col} \\ 0 & 0 & 0 & 0 \\ 0 & 0 & 0 & 0 \\ A_{lat} & A_{lon} & 0 & 0 \\ B_{lat} & B_{lon} & 0 & 0 \\ 0 & 0 & 0 & Z_{col} \\ 0 & 0 & N_{ped} & N_{col} \\ 0 & 0 & 0 & 0 \\ 0 & C_{lon} & 0 & 0 \\ D_{lat} & 0 & 0 & 0 \end{bmatrix} \begin{bmatrix} \delta_{lat} \\ \delta_{lon} \\ \delta_{ped} \\ \delta_{col} \end{bmatrix}$$

## II.  The model parameters of R-50 helicopter

The model parameters of the helicopter during hover and cruise-flight is summarized in the following table [5].

**Table 3: Control derivatives and time-constants of the Yamaha R-50**

| Parameter | Hover | Cruise |
|---|---|---|
| $B_{lat}$ | 0.14 | 0.124 |
| $B_{lon}$ | 0.0138 | 0.02 |
| $A_{lat}$ | 0.0313 | 0.0265 |
| $A_{lon}$ | -0.1 | -0.0837 |
| $Z_{col}$ | -45.8 | -60.3 |
| $M_{col}$ | 0 | 6.98 |
| $N_{col}$ | -3.33 | 0 |
| $N_{ped}$ | 33.1 | 26.4 |
| $D_{lat}$ | 0.273 | 0.29 |
| $C_{lon}$ | -0.259 | -0.225 |
| $Y_{ped}$ | 0 | 11.23 |
| $\tau_p$ | 0.0991 | 0.0589 |
| $\tau_f$ | 0.046 | 0.0346 |
| $h_{cg}$ | -0.411 | -0.321 |
| $\tau_s$ | 0.342 | 0.259 |